\documentclass[letterpaper, 10 pt, conference]{ieeeconf}  

\IEEEoverridecommandlockouts                              

\usepackage{graphicx}
\usepackage{amsmath}
\usepackage{amssymb}
\usepackage{booktabs}
\newcommand\pose{\textbf{LAMP}}

\usepackage[pagebackref,breaklinks,colorlinks]{hyperref}
\usepackage{paralist}
\usepackage{color}
\usepackage{xcolor}
\usepackage{hyperref}
\hypersetup{
    colorlinks=true,
    linkcolor=blue,
    filecolor=blue,      
    urlcolor=blue,
    citecolor=cyan,
}

\usepackage[capitalize]{cleveref}
\crefname{section}{Sec.}{Secs.}
\Crefname{section}{Section}{Sections}
\Crefname{table}{Table}{Tables}
\crefname{table}{Tab.}{Tabs.}

\graphicspath{{figs/}}

\title{\LARGE \bf
\pose: Leveraging Language Prompts for Multi-person Pose Estimation
}

\author{Shengnan Hu, Ce Zheng, Zixiang Zhou, Chen Chen, and Gita Sukthankar
\thanks{*This work was supported by Lockheed Martin Corporation.}
\thanks{All authors are with the University of Central Florida, Orlando, FL, USA}%
\thanks{Contact Email: shengnan.hu@ucf.edu} 
}

\begin{document}

\maketitle
\thispagestyle{empty}
\pagestyle{empty}

\begin{abstract}

Human-centric visual understanding is an important desideratum for effective human-robot interaction.  In order to navigate crowded public places, social robots must be able to interpret the activity of the surrounding humans.  This paper addresses one key aspect of human-centric visual understanding, multi-person pose estimation.  Achieving good performance on multi-person pose estimation in crowded scenes is difficult due to the challenges of occluded joints and instance separation. In order to tackle these challenges and overcome the limitations of image features in representing invisible body parts, we propose a novel prompt-based pose inference strategy called \pose\ (\textbf{L}anguage \textbf{A}ssisted \textbf{M}ulti-person \textbf{P}ose estimation). By utilizing the text representations generated by a well-trained language model (CLIP), \pose\ can facilitate the understanding of poses on the instance and joint levels, and learn more robust visual representations that are less susceptible to occlusion.  This paper demonstrates that language-supervised training boosts the performance of single-stage multi-person pose estimation, and both instance-level and joint-level prompts are valuable for training. The code is available at \url{https://github.com/shengnanh20/LAMP}. 
 
\end{abstract}

\section{Introduction}
\label{sec:intro}


Multi-person pose estimation (MPPE) algorithms must simultaneously solve two problems: detecting the people in the scene and localizing the joint keypoints. Dealing with occlusion is an important challenge for human pose estimation, especially in scenes with multiple persons and complex contexts.   
While existing algorithms have achieved promising results with both top-down and bottom-up methods, they often fail to recover occluded people and joints~\cite{zheng2020deep}.  Top-down methods~\cite{he2017mask,xiao2018simple,li2021pose,sun2019deep,yuan2021hrformer,ding20222r} usually rely on bounding-box cropping and cannot easily recover from instance detection failures or bounding-box overlap. Bottom-up methods~\cite{cao2017realtime,newell2017associative,jin2020differentiable,kreiss2019pifpaf,cheng2020higherhrnet,geng2021bottom,braso2021center,luo2021rethinking,wang2021robust,insafutdinov2016deepercut,pishchulin16cvpr, xue2022learning} experience difficulty estimating poses when there are overlapping individuals or occluded body parts, since they rely on grouping individual keypoints into poses. Also, bottom-up methods often require complex post-processing steps, which can be challenging to implement. Unlike top-down and bottom-up methods, the newer single-stage techniques~\cite{nie2019single,wei2020point,mao2021fcpose,shi2021inspose,zhou2019objects,shi2022end,wang2022contextual} have a simpler architecture and training process, which can reduce the risk of error accumulation. Since they don't rely on grouping individual keypoints, single-stage methods can be more efficient and typically employ dense regression to localize body keypoints directly from scene feature vectors. Due to the difficulty of long-range location regression, improving the accuracy of the regression procedure remains a challenge.

With the recent improvements of multi-modal training, language-supervised visual representation learning has shown great success on multiple tasks, including segmentation~\cite{luddecke2021prompt,xu2022}, depth estimation~\cite{zhang2022can} and action recognition~\cite{ju2022prompting}. CLIP (Contrastive Language-Image Pretraining) ~\cite{radford2021learning} is one of the most popular pre-training models, and we leverage it for our research.    Inspired by this line of research, this paper investigates the question: \emph{can language supervised training improve multi-person pose estimation by improving the detection of occluded figures and body keypoints?}  Our ultimate aim is to imbue a social robot with a human-centric visual understanding model that will enable it to navigate crowded streets, parks, and public places.  Our MPPE model can be used as an input for downstream tasks such as human mesh recovery~\cite{zheng2023feater,zheng2023potter}, action recognition~\cite{Hua2023SkeAttnCLR}, and person-to-person social interactions~\cite{zhao2023synthesizing}.

\begin{figure}
\centering
\includegraphics[width=0.5\textwidth]{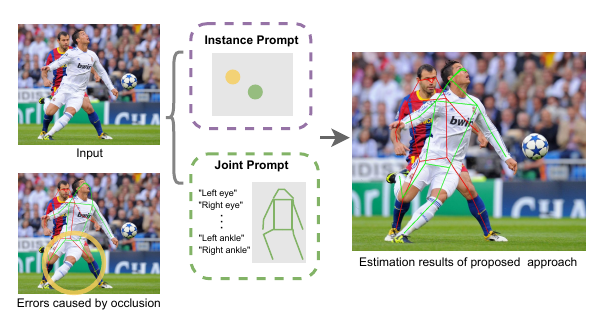}
\caption{Previous methods suffer from the difficulty of detecting overlapped persons and occluded keypoints. Our approach utilizes both instance and joint prompts.  The image model is trained to maximize the correlation between text and image features. The results on the occluded scenario (right) demonstrate the superiority of \pose.}

\label{fig:f1}
\end{figure}

\begin{figure*}
\centering
\includegraphics[width=\textwidth]{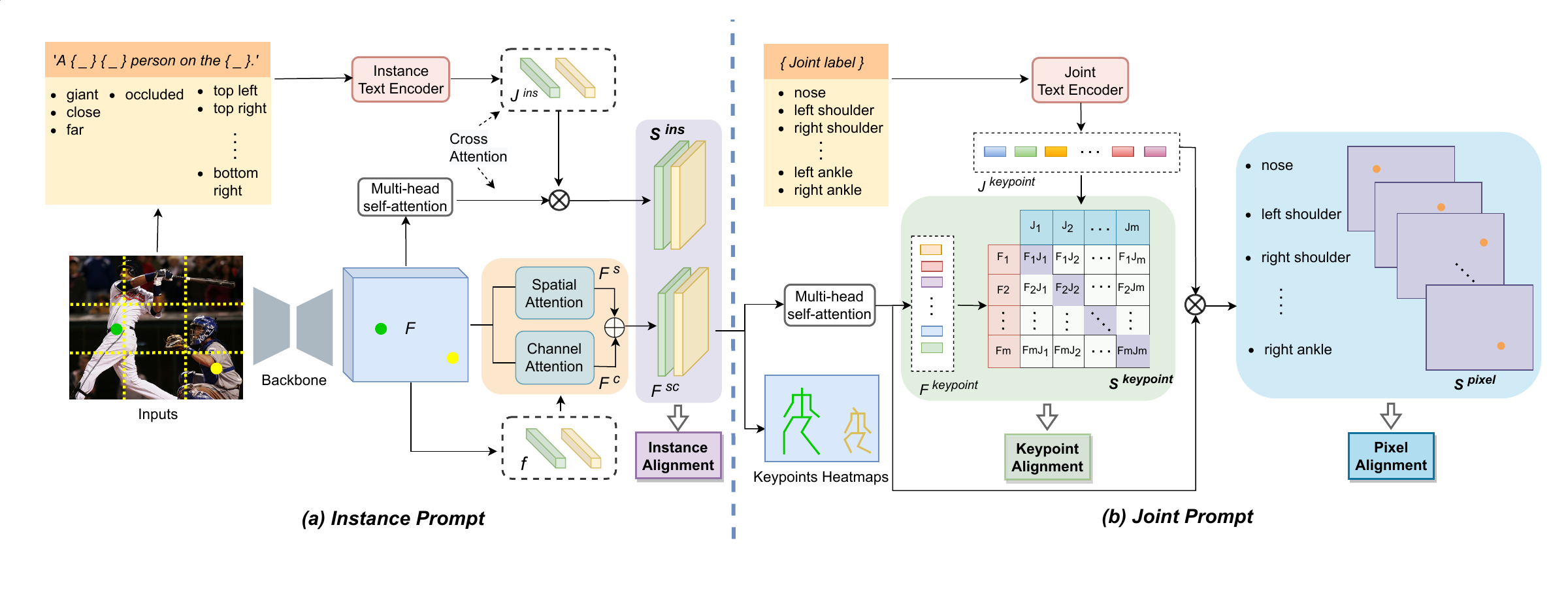}
\caption{Overview of the proposed architecture \pose.
(a) Given an input image, a backbone is applied to generate the global feature $F$. Then an image encoder with spatial attention and channel attention  extracts a set of instance-aware features $F^{sc}$ from the global features. A text encoder takes the description of each instance (Instance Prompt) as input and outputs a set of text features $J^{ins}$. A similarity-based textual embedding for each instance $S^{ins}$ is then generated by computing the inner product between the instance textual feature and the global image feature. (b) The instance-aware features are utilized by the Joint Prompt module which takes the joint labels as input and generates a list of textual features. Then a matching evaluation between image-text models for the joints is conducted on two levels: keypoint level and pixel level. \pose\ trains the image model to maximize the correlation between the textual embeddings and image features for the alignment modules shown in the figure.} 
\label{fig:intro}
\end{figure*}

Our proposed architecture (\pose) (\textbf{L}anguage \textbf{A}ssisted \textbf{M}ulti-person \textbf{P}ose estimation) uses text prompts to link semantic descriptions with both instances of people in the scene and joint keypoints, transforming the MPPE problem into a set of contrast tasks.   \pose\ utilizes both instance prompts and joint prompts and trains the image model to maximize the correlation between text and image features.  Prompts are generated automatically with templates and are only used during the supervised training phase.
\emph{Instance prompts} describe the relative locations of the people in the scene and whether or not they are occluded.  This information is used to refine instance-aware heatmaps and helps \pose\ ignore extraneous people in the scene.  
The instance-aware heatmaps are then used for joint localization. \emph{Joint prompts} contain text describing the body keypoint (e.g. nose, left ankle, right shoulder).  The CLIP model is trained on sufficient data to enable these joint descriptions to be semantically meaningful.  To improve joint localization, we incorporate both keypoint-level alignment and pixel-level alignment.  The intuition is that the inclusion of language pre-training makes MPPE more robust to occlusion, while also improving the performance of pixel-level alignment.  Figure~\ref{fig:intro} shows the \pose\ architecture.


Although CLIP is a tremendously valuable tool, adapting it to perform well on downstream applications such as multi-person pose estimation is not straightforward.
This paper evaluates the performance of \pose\ on OCHuman~\cite{zhang2019pose2seg} and CrowdPose~\cite{li2019crowdpose}.  \pose\ outperforms the other single-stage and bottom-up methods and is competitive against the best top-down methods.  We also perform an ablative study to show the relative contributions of the instance and joint components of the vision-language model on \pose's performance.

This paper makes the following contributions:
\begin{compactitem}
\item introduces a new architecture for multi-person pose estimation for social robots that incorporates language-assisted visual representation learning;
\item demonstrates the power of language for making the image model robust to occlusion and improving the accuracy of the dense regression;
\item outperforms existing single-stage models for MPPE on two challenging benchmarks for crowded pose estimation: OCHuman and CrowdPose. 
\end{compactitem}




\section{Related Work}
\label{sec:related work}

\subsection{Multi-person Pose Estimation}
Multi-person pose estimation (MPPE) comprises two problems: 1)  detecting all instances of people in the scene and 2) localizing their body keypoints.   MPPE methods need to overcome both occlusion created by crowded scenes, as well as self-occlusion, where one body part obscures another.  A secondary problem is handling scale and orientation diversity between people in the scene; it is difficult to perform human pose estimation on figures that may be only a few pixels in height.   
\subsubsection{Top-down}
Top-down methods~\cite{he2017mask,xiao2018simple,li2021pose,sun2019deep,yuan2021hrformer,ding20222r} typically identify a bounding box for each person in the scene and then perform single person pose estimation. This two stage process often makes top down detectors computationally slower but gives them an edge on average precision.   Mask-RCNN~\cite{he2017mask} reuses feature maps from the detector for pose estimation through the RoIalign operation, making it more efficient than other top down methods.  HRNet~\cite{sun2019deep} seeks to improve pose estimation performance by maintaining high resolution representations throughout the whole process, rather than recovering  them from a high-to-low resolution network.  We use the HRNet~\cite{sun2019deep} as a backbone for several of the benchmarks.   Xiao et al.~\cite{xiao2018simple} published a comprehensive evaluation of the effect of different detectors, backbone choices, and pose estimation algorithms on top down MPPE performance, and we benchmark \pose\ vs.\ their simple baselines in addition to other SOTA algorithms.

\subsubsection{Bottom-up} 
Bottom-up algorithms~\cite{cao2017realtime,newell2017associative,jin2020differentiable,kreiss2019pifpaf,cheng2020higherhrnet,geng2021bottom,braso2021center,luo2021rethinking,wang2021robust,pishchulin16cvpr,insafutdinov2016deepercut} group identity-free keypoints into persons using a variety of methods; for instance, DeepCut~\cite{pishchulin16cvpr,insafutdinov2016deepercut} formulates the grouping problem as an integer linear program.  Other grouping methods include part linking~\cite{cao2017realtime,kreiss2019pifpaf}, associative embedding~\cite{newell2017associative}, and hierarchical clustering~\cite{jin2020differentiable}.
Separate heatmaps are usually calculated for each keypoint; a strength of these methods is that the heatmaps often outperform direct pixel-wise regression at localizing keypoints~\cite{sun2020bottom}. 
In general, bottom-up methods are more computationally efficient than top-down techniques due to their simpler convolutional pipelines but exhibit slightly inferior performance.   


\subsubsection{Single-stage}
Single-stage approaches~\cite{nie2019single,wei2020point,mao2021fcpose,shi2021inspose,zhou2019objects,shi2022end,wang2022contextual} densely regress candidate body keypoints directly from the original image, creating an end-to-end trainable pipeline.   These methods are better at preserving valuable contextual information from the scene that gets discarded by both top-down and bottom-up systems.  Also they do not rely on specific modules to perform RoI cropping, non-maximal suppression, and keypoint grouping.  A key concern for these methods is improving the accuracy of the regression procedure.  
Contextual Instance Decoupling (CID)~\cite{wang2022contextual} improves the quality of regression in MPPE by separating the multi-person heatmap into instance-aware heatmaps that are later used to estimate heatmaps and keypoints for each person in the scene.  We focus on further improving the performance of single-stage methods through the additional vision-language pretraining.

\subsection{Vision-Language Models}
Language has been shown to facilitate the learning of effective visual representations for zero-shot domain transfer on downstream tasks. CLIP (Contrastive Language-Image Pre-training) \cite{radford2021learning}, a large-scale language-vision model, trains a textual encoder and a visual encoder on a dataset of 400M image-text pairs to explore the relationship between the modalities.   Usually CLIP~\cite{radford2021learning} is provided with a text prompt engineered for the downstream task to create a contrast task.  The inclusion of the prompt during model training makes the downstream task more similar to the task originally used to create to model. Liu et al.~\cite{liuprompt2021}  present a survey of prompt creation approaches.  
Although many of the downstream tasks that have benefited from CLIP~\cite{radford2021learning} are simpler detection and classification tasks, DepthCLIP~\cite{zhang2022can} illustrates that it is possible to leverage CLIP for pixel level tasks.  PromptPose~\cite{zhang2022promptpose} uses vision-language pre-training for animal pose estimation; they also use prompt and pixel-level loss functions for body keypoint localization but do not learn instance level models.
\section{Method}
\label{sec:method}
Our work leverages the text representations generated by language models such as CLIP \cite{radford2021learning} to facilitate the understanding of poses. Towards this end, we conduct a comprehensive study on the design and generation of pose prompts including instance level information and joint level information, which empowers the pose estimation baseline model in scenarios with multiple people.

\subsection{Overview}
Vision-language models have achieved recent successes in enhancing the performance of many vision tasks, which shows promising transfer capability between vision and text domains. Specifically, given a paired image and text $(image, text)$, vision-language models learn a multi-modal embedding space by jointly training an image encoder and text encoder to maximize the cosine similarity between the correct pair while minimizing the incorrect pairs \cite{radford2021learning}. Using a trained embedding, vision-language models can achieve zero-shot transfer to downstream image recognition tasks. Therefore, inspired by the previous work on introducing language supervision into visual recognition tasks \cite{zhang2022can, zhang2022promptpose}, we introduce a learned prior and investigate the use of text prompts to enhance the recognition of instances and keypoints in multi-person pose estimation.



Figure. \ref{fig:intro} shows the flow of our instance-joint prompt pipeline for multi-person pose estimation. There are two key modules in the proposed architecture: instance prompt and joint prompt, where the instance prompt is used to provide dense instance-related information and semantics, and the joint prompt is used to refine the multi-modal connections at a joint level.

Given an image $I\in \mathbb{R}^{3 \times H \times W}$, we extract multi-scale image features from the backbone (i.e. HRNet \cite{sun2019deep}). The multi-scale features are then concatenated together and generate the global image feature map $F$. After that, we employ the instance decoupling module \cite{wang2022contextual} with spatial attention and channel attention as the image encoder, to obtain the image features for each instance. An instance prompt with a pretrained language-image encoder (i.e. CLIP) is simultaneously used to exploit the instance-related features in the textual model thus providing effective supervision to improve the discriminative power of the instance encoder. 

Instance-aware feature maps are then fed to the heatmap module with a CNN layer, to obtain the probability distribution maps for the keypoints, i.e. keypoint heatmaps. 

In order to leverage the semantic information from the language model for accurate keypoint estimation, we propose a method that incorporates both keypoint-level and pixel-level contrast. This approach establishes the connections between the image features and the embedding prompts, allowing for a more effective utilization of joint-related semantics in the language model.


With the alignment training by the proposed model, we can align the instance and joint features between the image model and language model, thus providing an effective prompt to guide the image module to learn a powerful visual representation in the MPPE task.

\subsection{Instance Prompt with Attribute Injection}
\label{method_ins}
\subsubsection{Image Encoder}
Following previous studies of attention mechanism \cite{hu2018squeeze, woo2018cbam, wang2022contextual,Zhao_2023_CVPR}, we decouple the cues of each instance from the original feature map based on spatial attention and channel attention. Specifically, given a feature map $F$, we first roughly extract the features of instances based on the ground truth locations of their center points, obtaining a set of instance features $f \in \mathbb{R}^{N \times C^ {ori}}$. After that, a spatial attention module is used to separate instances into different spatial locations, while a channel attention module is used to decouple instances into different channels of the feature map. 

To obtain the spatial feature of each instance, we compute an instance-aware mask $M^i$ based on the location of the $i$-th instance following \cite{wang2022contextual}. Then we generate the spatial recalibration as:
\begin{equation}\label{1}
F^{s}_i = F \cdot M_i.
\end{equation}

In order to decouple the channel feature map of the $i$-th instance, we perform an element-wise manipulation between the instance feature $f_i$ and the global feature as follows:
\begin{equation}\label{2}
F^{c}_i = F \otimes  f_i, 
\end{equation}
where $f_i$ is generated from the original feature based on features at the coordinates of the instance's center point.

After obtaining the spatial recalibration and channel recalibration of each instance, we then conduct a fusion recalibration to bolster the discriminative ability of the model among different instances. Thus the fused instance-aware features are calculated as:
\begin{equation}\label{3}
F^{sc} = Relu(Conv([F^{s}_i; F^{c}_i]))  \in \mathbb{R}^{N \times C \times H \times W}.
\end{equation}

\subsubsection{Textual Label Generation}
At the core of our approach is the idea of prompting person pose perception from supervision contained in the language. Inspired by the findings in \cite{shen2022k, qin2022medical} that an expressive description can  benefit the transfer performance of vision language models, here we design a prompt for instance separation by injecting essential attributes of instances into the model. 

The detailed structure of the instance-level prompt is illustrated in Figure \ref{fig:intro} (a). For each instance $instance_i$ in the original image, we construct the prompt by applying the following template: 
\begin{equation}\label{4}
\begin{split}
Prompt_{i} = Template(\left\{s^{Attr_{m}} \right\}, instance_i), \\ 
Attr_{m} \in \left\{ location, depth, occlusion \right\}.
 \end{split}
\end{equation}

Specifically, to calculate the location information of the instance, we divide the original image into $3 \times 3$ patches and label the instance using the coordinates of the center point of each instance's bounding box. To enhance the discrimination between different instances, we introduce a pseudo depth label based on the scale of each instance's bounding box. This is achieved by computing the ratio of the width and height of the bounding box to the width and height of the image. Based on this ratio, we assign approximate depth labels, such as "far," "close," and "giant," to the different instances within the image. This allows for a rough categorization of depth information based on the relative size of the instances. Another attribute that can distinguish the relationships between instances further is occlusion. By counting the visible keypoints of each instance, we can decide whether an instance is occluded, which can also help establish the depth orders of instances. Finally, prompts formed by instance attributes will pass through the textual encoder of CLIP \cite{radford2021learning} into latent space by:
\begin{equation}\label{5}
P^{ins} = TextualEncoder(Prompt_{i}) \in \mathbb{R}^{N \times C_{emb}}.
\end{equation}


A transformer decoder layer with a multi-head self-attention mechanism is then applied to enhance the projection of the textual feature to the image feature, and we have the textual prompts $J^{ins}_i \in \mathbb{R}^{1 \times C_{emb}}$ for the $i$-th instance.

\subsubsection{Instance-level Alignment}
\label{align}
To explore the connections between the image feature and text embedded feature of each instance, it is necessary to project both of them into a multi-modal embedding space. Here for the image feature, we conduct a multi-head self-attention layer with a feed-forward network (FFN) after the original feature to get the normalized image encoded feature $F^{img} \in \mathbb{R}^{N \times C \times H \times W}$.

We then calculate the inner product between the mapped image features and embedded textual features $J^{ins}$ and get an instance-pixel matching score map of each instance:
\begin{equation}\label{6}
S^{ins}_i = F^{img} \cdot J^{ins}_i \in \mathbb{R}^{H \times W}.
\end{equation}

With such a similarity-based ensemble, we could fuse the prior knowledge learned by CLIP \cite{radford2021learning} and the image feature generated by the backbone and achieve a more informative representation for each instance. Thus by conducting an alignment between the encoded image features $F^{sc}$ and the embedded textual features $S^{ins}$, \pose\ provides extra supervision and boosts the discriminative power of the instance image encoders. Finally, an MSE loss between the textual embeddings and the image feature is calculated to maximize the correlation between them.


\subsection{Joint Prompt}

As shown in Figure \ref{fig:intro} (b), after obtaining the instance-aware image features, the joint-level prompt is designed to further refine the pose representation at a joint level. Inspired by \cite{zhang2022promptpose}, we enhance the textual prompt by utilizing the embeddings from two aspects: keypoint level and pixel level.

\subsubsection{Keypoint-level Alignment}
\label{keypoint}

As mentioned in section \ref{align}, the image features of instances $F^{sc}$ are first fed into a multi-head self-attention layer to map into the multi-modal embedding space, obtaining $F^{ins} \in \mathbb{R}^{N \times C \times H \times W}$. 

Then to establish the connections between the text and local keypoint features in the image, a keypoint-level alignment is conducted by exploring the feature similarities in the embedding domain. Concretely, given $m$ joints, the textual encoder takes the description word of each body joint (e.g. nose, left shoulder, right shoulder) as input and then generates the textual features of the joints $J^{keypoint} \in \mathbb{R}^{m \times C_{emb}}$, where $m$ is the number of joints.

To align the textual joint features with the local feature of the joints in the image, we utilize the ground truth locations of the joints to sample features from the original instance image features, obtaining keypoint local features $F^{keypoint} \in \mathbb{R}^{m \times C_{emb}}$. Then we calculate the cosine similarities between each $\left \{ image,  text \right \}$ pair of the joints and obtain the similarity map as:

\begin{equation}\label{7}
S^{keypoint}_i = F^{keypoint}_i \cdot (J^{keypoint}) ^T \in \mathbb{R}^{m \times m}.
\end{equation}

We then perform contrastive learning on the keypoint local features and textual embedding features based on the contrastive loss.



\subsubsection{Pixel-level Alignment}

Furthermore, to establish a dense connection between the image features and the joint textual features in the embedding space, a pixel-level prompt is introduced to enhance the keypoints regression.

Following section \ref{keypoint}, we obtain the textual embedding features of the joints  $J^{keypoint} \in \mathbb{R}^{m \times C_{emb}}$. Here, we apply an inner product between the instance image features $F^{ins}$ and the joint textual features in the embedding space and get the text-pixel matching score of each pixel location by:

\begin{equation}\label{8}
S^{pixel}_i = F^{ins}_i \cdot  J^{keypoint} \in \mathbb{R}^{H \times W}.
\end{equation}

This score map is supervised by the target heatmaps through the pixel-level contrastive loss.

\subsection{Loss Functions}

\subsubsection{Image Loss}
Similar to \cite{wang2022contextual}, we train the instance decoupling attention model with a contrastive loss to enhance the discriminative power of each instance. Given a set of instance features $f \in \mathbb{R}^{N \times C^ {ori}}$, we obtain:
\begin{equation}\label{9}
L_{con}^{f_i} = -log  \frac{exp(\left\|f_i \right\|^2 /\tau)}{\sum_{n}^{j=1}  exp(\left\|f_i \right\| \left\|f_j \right\| / \tau)},
\end{equation}

where $n$ is the total count of instances in the image, and $\tau$ is a temperature parameter, which is set to 0.5 in experiments.

For heatmap regression, we compute the Focal Loss \cite{law2018cornernet} between the predicted heatmaps and ground truth heatmaps generated based on the keypoints, obtaining:

\begin{equation}\label{10}
\resizebox{1.0\hsize}{!}{
$\begin{aligned}
FL(p, h^{gt}) = & \frac{-1}{N}\sum_{c=1}^{C}\sum_{x=1}^{H}\sum_{y=1}^{W} \\
&\begin{cases}
   (1-p_{cxy})^{\alpha}log(p_{cxy}) &\text{ if }  h_{cxy}=1 \\
   (1-h_{cxy})^{\beta} (h_{cxy})^{\alpha}log(1-h_{cxy}))   &\text{otherwise},
\end{cases}
\end{aligned}$
}%
\end{equation}

where $h$ denotes the ground truth heatmaps and $p$ denotes the predicted heatmaps. Thus we have the overall heatmap loss as:
\begin{equation}\label{11}
L_{FL}^{f} = \frac{1}{N}\sum_{i=1}^{N} FL(p_i, h^{gt}_i).
\end{equation}

\subsubsection{Prompt Loss}
After obtaining the instance textual embeddings, an MSE loss is computed between the textual embeddings and the image features of each instance, i.e.,
\begin{equation}\label{12}
L_{p}^{ins} = MSE(S^{ins}, F^{sc} )
\end{equation}

As for keypoint-level alignment, we perform a contrastive learning by computing the cross entropy classification loss between the obtained similarity map $S^{keypoint} \in \mathbb{R}^{m \times m}$ and the matching target $M_{gt} = [0, 1, 2 ..., m-1]$ as:
\begin{equation}\label{13}
\resizebox{1.0\hsize}{!}{$
L_{p}^{keypoint} = \frac{1}{2}(CE(S^{keypoint}, M_{gt}) + CE((S^{keypoint})^T, M_{gt})
$}%
\end{equation}

To enable effective joint-pixel embedding, we utilize the ground truth heatmaps as the supervision of the joint-pixel matching map by computing the MSE loss as:
\begin{equation}\label{14}
L_{p}^{pixel} = MSE(S^{pixel}, h^{gt})
\end{equation}

 Formally, the overall loss function of our approach can be formulated as:
\begin{equation}\label{15}
L = \lambda_{1} L_{con}^{f} + \lambda_{2} L_{FL}^{f} + \lambda_{3} L_{p}^{ins} + \lambda_{4} L_{p}^{keypoint} +  \lambda_{5} L_{p}^{pixel}, 
\end{equation}
where $\lambda_{1-5}$ are the loss weights, respectively.  

\section{Experiments}
\subsection{Datasets and Metrics}
We follow the protocol of the recent state-of-the-art multi-person pose estimation method (i.e. CID \cite{wang2022contextual}) and evaluate the performance of the proposed method on two multi-person pose estimation benchmarks: OCHuman \cite{zhang2019pose2seg} and CrowdPose \cite{li2019crowdpose}.


 \paragraph{OCHuman} This is a challenging dataset with many crowded scenes.  It contains 4731 images with 8110 persons in total, including 2500 for training and validation, and 2331 for testing. We report our results following the setting of previous methods \cite{wang2022contextual} \cite{qiu2020peeking}.  
 
 \paragraph{CrowdPose} This is another dataset with highly occluded scenarios. The dataset contains $10K$ images and $80K$ annotated person instances with 14 keypoints for training, $2K$ images for validation and $8K$ images for testing.

 \paragraph{Evaluation Metric} We follow the standard evaluation metric~\cite{zheng2020deep} and use the Object Keypoint Similarity (OKS) metrics for pose estimation. We report average precision and average recall scores with different thresholds and different object sizes: $AP, AP^{50}, AP^{75}, AP^{M}, AP^{L}, AR,  AR^{M}$ and $AR^{L}$.

\subsection{Implementation Details}
\pose\ was implemented using Pytorch \cite{paszke2017automatic} and MMPose \cite{mmpose2020}.  Following previous work \cite{geng2021bottom, wang2022contextual}, we use the HRNet-W32 \cite{sun2019deep} pretrained on ImageNet as the backbone for the image encoder in the pose estimation pipeline. For the language model, we use the text encoder of CLIP-ViT-Base \cite{radford2021learning} and initialize the model with the corresponding pre-trained weights. 

During training, the input images are resized to $512 \times 512$, followed by random rotation, scale jitter, and translation. We use the Adam optimizer \cite{kingma2014adam} and set the initial learning rate as 0.001. Note that text encoders are frozen during training to prevent the language knowledge from being distracted \cite{zhang2022promptpose}.

For testing, images are resized with the short side to 512 while maintaining the aspect ratio between height and width. 

\begin{figure*}
\centering
\includegraphics[width=\textwidth]{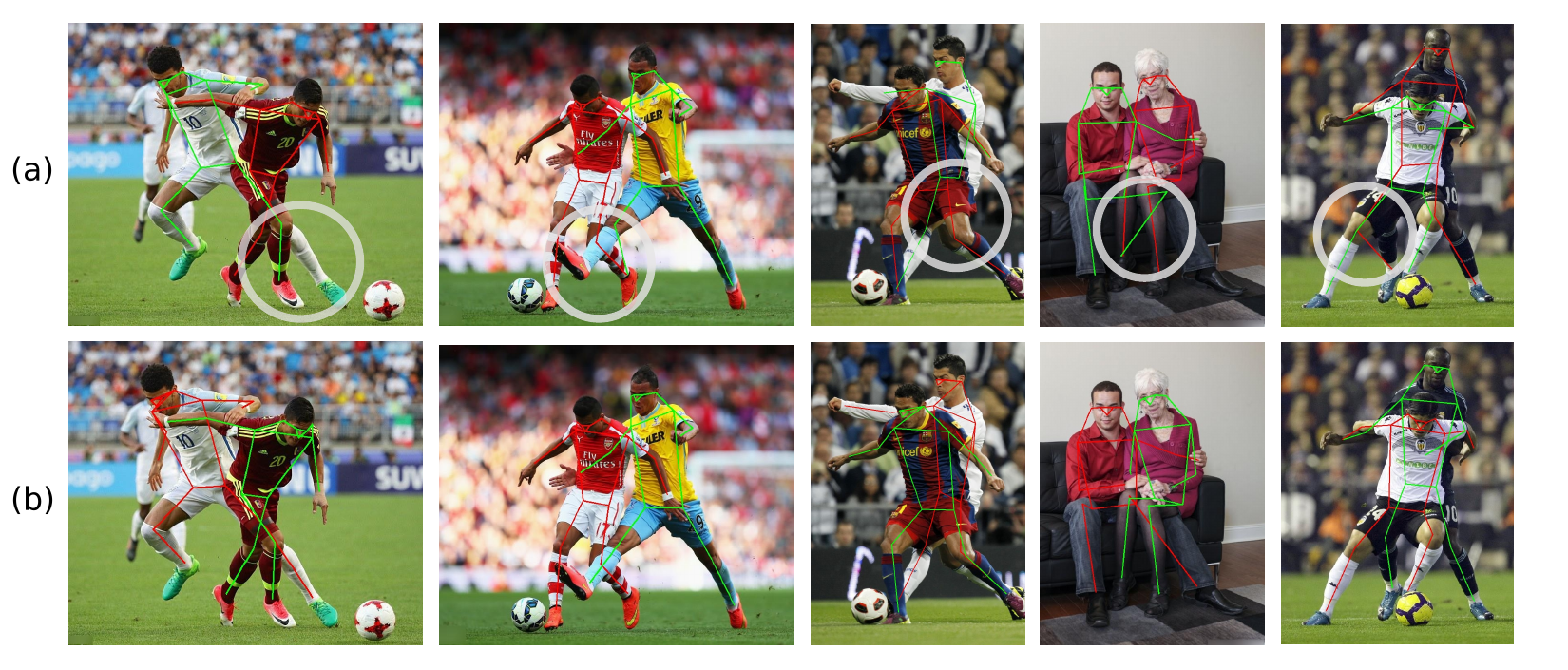}
\caption{Qualitative comparison between the proposed method \pose\ and the state-of-the-art approach CID on OCHuman testing set. (a) Estimation results produced by CID. (b) Estimation results produced by \pose. Compared with CID, our proposed method \pose\ is more robust to person overlapping and occlusion.}
\label{ fig:vis}
\end{figure*}

\begin{table}[]
\centering
\resizebox{\columnwidth}{!}{%
\begin{tabular}{lccc}
\hline
\multicolumn{1}{l|}{Method}            & $AP$$\uparrow$           & $AP^{50}$$\uparrow$        & $AP^{75}$$\uparrow$        \\ \hline
\multicolumn{4}{c}{Top-down methods}                                                   \\ \hline
\multicolumn{1}{l|}{Mask R-CNN \cite{he2017mask}}        & 20.2          & 33.2          & 24.5          \\
\multicolumn{1}{l|}{SimplePose \cite{xiao2018simple}}        & 24.1          & 37.4          & 26.8          \\
\multicolumn{1}{l|}{OPEC-Net~\cite{qiu2020peeking}}        & 32.8       &60.5      &31.1          \\

\hline
\multicolumn{4}{c}{Bottom-up methods}                                                  \\ \hline
\multicolumn{1}{l|}{HigherHRNet \cite{cheng2020higherhrnet}}       & 27.7          & 66.9          & 15.9          \\
\multicolumn{1}{l|}{HGG \cite{jin2020differentiable}}             & 36.0          & -          & -          \\ 

\hline
\multicolumn{4}{c}{Single-stage methods}                                               \\ \hline
\multicolumn{1}{l|}{SPM \cite{nie2019single}}               & 47.6          & 67.5          & 53.2          \\
\multicolumn{1}{l|}{DEKR (HRNet-W32) \cite{geng2021bottom}}   & 52.2          & 69.9          & 56.6          \\
\multicolumn{1}{l|}{CID (HRNet-W32) \cite{wang2022contextual}}    & 57.5          & 75.5          & 63.3          \\ \hline
\multicolumn{1}{l|}{Ours(HRNet-W32)}   & \textbf{58.8} & \textbf{76.6} & \textbf{64.3}  \\ \hline
\end{tabular}%
}%
\caption{Comparisons with state-of-the-art methods on the OCHuman testing set.}
\label{tab:ochuman}
\end{table}

\begin{table*}[]
\centering
\begin{tabular}{lcccccccc}
\hline
\multicolumn{1}{l|}{Method}                            & \multicolumn{1}{c|}{Backbone}   & \multicolumn{1}{c|}{Input size} & $AP$$\uparrow$   & $AP^{50}$$\uparrow$  & \multicolumn{1}{c|}{$AP^{75}$$\uparrow$} & $AP^{E}$$\uparrow$ & $AP^{M}$$\uparrow$ & $AP^{H}$$\uparrow$ \\ \hline
\multicolumn{9}{l}{Top-down methods}                                                                                                                                                                                                                                                                  \\ \hline
\multicolumn{1}{l|}{Mask-R-CNN \cite{he2017mask}}                        & \multicolumn{1}{c|}{ResNet-50}  & \multicolumn{1}{c|}{800}        & 57.2 & 83.5                   & \multicolumn{1}{c|}{60.3}                   & 69.4                  & 57.9                                       & 45.8                  \\
\multicolumn{1}{l|}{SimplePose \cite{xiao2018simple}} & \multicolumn{1}{c|}{ResNet-512} & \multicolumn{1}{c|}{384x288}    & 60.8 & 81.4                   & \multicolumn{1}{c|}{65.7}                   & 71.4                  & 61.2                                       & 51.2                  \\
\multicolumn{1}{l|}{AlphaPose \cite{fang2017rmpe}}                         & \multicolumn{1}{c|}{}           & \multicolumn{1}{c|}{-}          & 61.0 & 81.3                   & \multicolumn{1}{c|}{66.0}                   & 71.2                  & 61.4                                       & 51.1                  \\
\hline
\multicolumn{9}{l}{Bottom-up methods}                                                                                                                                                                                                                                                                 \\ \hline
\multicolumn{1}{l|}{OpenPose \cite{cao2017realtime}}                          & \multicolumn{1}{c|}{VGG-19}     & \multicolumn{1}{c|}{-}          & -    & -                      & \multicolumn{1}{c|}{-}                      & 62.7                  & 58.7                                       & 32.3                  \\
\multicolumn{1}{l|}{HigherHRNet \cite{cheng2020higherhrnet}}                           & \multicolumn{1}{c|}{HRNet-W32}  & \multicolumn{1}{c|}{512}        & 65.9 & 86.4                   & \multicolumn{1}{c|}{70.6}                   & 73.3                  & 66.5                                       & 57.9                  \\

\multicolumn{1}{l|}{PINet \cite{wang2021robust}}                            & \multicolumn{1}{c|}{HRNet-W32}  & \multicolumn{1}{c|}{512}        & 68.9 & 88.7                   & \multicolumn{1}{c|}{74.7}                   & 75.4                  & 69.6                                       & 61.5                  \\ \hline
\multicolumn{9}{l}{Single-stage methods}                                                                                                                                                                                                                                                              \\ \hline
\multicolumn{1}{l|}{SPM \cite{nie2019single}}                               & \multicolumn{1}{c|}{Hourglass}  & \multicolumn{1}{c|}{-}          & 63.7 & 85.9                   & \multicolumn{1}{c|}{68.7}                   & 70.3                  & 64.5                                       & 55.7                  \\
\multicolumn{1}{l|}{DEKR(HRNet-W32) \cite{geng2021bottom}}                   & \multicolumn{1}{c|}{HRNet-W32}  & \multicolumn{1}{c|}{512}        & 65.7 & 85.7                   & \multicolumn{1}{c|}{70.4}                   & 73.0                  & 66.4                                       & 57.5                  \\
\multicolumn{1}{l|}{CID(HRNet-W32) \cite{wang2022contextual}}                    & \multicolumn{1}{c|}{HRNet-W32}  & \multicolumn{1}{c|}{512}        & 71.2 & 89.8                   & \multicolumn{1}{c|}{76.7}                   & 77.9                  & 71.9                                       & 63.8                  \\ 
\multicolumn{1}{l|}{E2Pose \cite{tobeta2022e2pose}}                    & \multicolumn{1}{c|}{ResNet-512}  & \multicolumn{1}{c|}{512}        & 66.5 & -                   & \multicolumn{1}{c|}{-}                   & -                  & -                                       & -                 \\ 

\hline
\multicolumn{1}{l|}{Ours}                              & \multicolumn{1}{c|}{HRNet-W32}  & \multicolumn{1}{c|}{512}        & \textbf{71.4} & \textbf{90.3}                       & \multicolumn{1}{c|}{\textbf{77.1}}                       &\textbf{77.9}                       & \textbf{72.1}                                           & \textbf{64.2}                      \\ \hline
\end{tabular}
\caption{Comparison with state-of-the-art methods on CrowdPose testing set.}
\label{tab:crowdpose}
\end{table*}

\begin{table}[]
\centering
\resizebox{\columnwidth}{!}{%
\begin{tabular}{l|ccc}
\hline
Model                           & $AP$$\uparrow$            & $AP^{50}$$\uparrow$ & $AP^{75}$$\uparrow$ \\ \hline
Base (Image only)                            & 57.5          & 75.5                   & 63.3                   \\
Base + Instance Alignment                 & 57.9          & 75.9                   & 63.4                   \\
Base + Keypoint Alignment                   & 57.8          & 76.2                   & 63.8                   \\
Base + Pixel Alignment                   & 58.6          & 76.5                   & 63.7                   \\
Base + Instance + Keypoint + Pixel & \textbf{58.8} & \textbf{76.6}          & \textbf{64.3}          \\ \hline
\end{tabular}
}%
\caption{Ablation study on OCHuman.}
\label{tab:ablation}
\end{table}

\subsection{Comparison with SOTA}

\paragraph{Results on OCHuman Testing Set} To assess the performance of \pose\ in tackling occlusions, we compare it with state-of-the-art methods on the OCHuman dataset, one of the most challenging benchmarks for crowded pose estimation. This dataset presents a significant challenge due to increased person overlapping and occlusions, which can negatively impact person detection and keypoint localization. As depicted in Table \ref{tab:ochuman}, we compare three categories of methods: top-down (Mask R-CNN, SimplePose, and OPEC-Net), bottom-up (HigherHRNet and HGG), and single-stage (SPM, DEKR, and CID). Our results show that \pose\ outperforms all top-down methods, demonstrating the superiority of our decoupling strategy in crowded scenes. It also outperforms bottom-up methods by achieving 22.8\% AP higher than HGG. Additionally, \pose\ surpasses single-stage methods, with a 1.3\% improvement over the most recent work CID.

\paragraph{Results on CrowdPose Testing Set} Table~\ref{tab:crowdpose} presents the comparison of \pose\ on another challenging benchmark CrowdPose. It can be observed that \pose\ achieves the best performance, outperforming CID by 0.2\%. A noteworthy finding is that the proposed model achieves a higher improvement over previous works on $AP^{H}$, outperforming CID by 0.4\%. This further supports the efficacy of \pose\ in handling heavy occlusions and overlapping people.

\paragraph{Qualitative Results} Fig. \ref{ fig:vis} visualizes some pose estimation results produced by the proposed model on the OCHuman testing set (bottom). This visual representation showcases the ability of our model to accurately estimate multiple poses in crowd scenarios, particularly for overlapped persons and occluded body parts.

\subsection{Ablation Study}
In this section, we perform an ablation study to investigate the contribution of each prompt component in the proposed model. The results are presented in Table \ref{tab:ablation}. We can observe that, by introducing the instance alignment and the keypoint alignment, the proposed approach improves the performance to 57.9\% AP and 57.8\% AP respectively. This validates the effectiveness of the instance prompt and joint prompt. Notably, the model obtains the largest performance improvement with the help of pixel alignment, and the complete \pose\ further outperforms the baseline and achieves SOTA performance on the OCHuman dataset. This can be attributed to pixel-level alignment that maximizes the correlation between text prompts and image features more precisely, making the model more robust in the presence of overlapping persons or invisible joints. These results further demonstrate the superiority of the proposed \pose\ architecture in occluded scenarios.

\section{Conclusion and Future Work}
Developing models of human-centered visual understanding that are robust to occlusion is an important stepping stone towards the development of social robots that can navigate crowded public places.
In this paper, we tackle the challenge of multi-person pose estimation in crowded scenes. Our proposed \pose\ architecture exhibits substantial improvements in occluded pose estimation through its end-to-end pipeline that leverages both instance and joint cues from the language model. Results from experiments on several crowded pose estimation benchmarks confirm the efficacy of our proposed method in addressing occlusion. 

In future work, we will extend our model to handle downstream tasks such as detecting person-to-person social interactions and recognizing non-verbal communication.  We envision our model as being useful for the human-robot interaction components of the Robocup@Home competition, assisting navigation and user control in UAV-human workspaces, and facilitating trust in human-robot teaming.

\bibliographystyle{IEEEtran} 
\bibliography{IEEEabrv,mybib}



\end{document}